\documentclass[conference]{IEEEtran}
\IEEEoverridecommandlockouts
\usepackage{cite}
\usepackage{amsmath,amssymb,amsfonts}
\usepackage{algorithmic}
\usepackage{graphicx}
\usepackage{textcomp}
\usepackage{xcolor}
\def\BibTeX{{\rm B\kern-.05em{\sc i\kern-.025em b}\kern-.08em
    T\kern-.1667em\lower.7ex\hbox{E}\kern-.125emX}}

\usepackage[
    letterpaper,
    left=0.75in,
    right=0.75in,
    top=0.75in,
    bottom=0.78in
]{geometry}

\begin{document}

\newgeometry{
    top=1in,
    bottom=0.78in,
    left=0.75in,
    right=0.75in
}

\title{A Systematic Evaluation of Infrastructure-Based Radar System for Highway Traffic Monitoring}

\author{
\IEEEauthorblockN{
Tianheng Zhu,
Woei-chyi Chang,
Alamss Riaz,
Sogand Hasanzadeh,
Yiheng Feng
}
\IEEEauthorblockA{
\textit{Lyles School of Civil and Construction Engineering} \\
\textit{Purdue University}, West Lafayette, IN, USA \\
\{zhu1230, chang803, riaz7, sogandm, feng333\}@purdue.edu
}
}

\maketitle

\begin{abstract}
Infrastructure-based radar systems offer robust and long-range solutions for traffic monitoring, yet their detection and tracking performance under real-world conditions remains insufficiently evaluated. This study introduces DRaT (Drone and Radar Trajectories), a dual-modality dataset of naturalistic vehicle trajectories collected at a highway merging segment in Fort Worth, Texas, to systematically assess radar sensing performance against drone-derived ground truth. The performance is evaluated at three levels: individual vehicle detection, trajectory tracking, and macroscopic traffic parameter estimation. For individual vehicle detection, the radar achieves an overall precision of 78\% and a recall of 57\%, with degraded performance under congested traffic conditions and at longer distances. At the trajectory level, the radar demonstrates reasonably strong tracking performance (IDF1 $=$ 0.699), maintaining reliable vehicle identities when tracks are successfully established. For macroscopic traffic flow metrics, the radar accurately estimates space-mean speed (MAPE $<$ 4\%) but underestimates density and volume by approximately 23\% due to missed detections. The paper also discusses practical deployment considerations and potential downstream applications of roadside radar sensing systems. To support reproducible research on infrastructure-based sensing systems, we have open-sourced the DRaT dataset on Zenodo: https://zenodo.org/records/20171110.
\end{abstract}


\section{Introduction}
High-resolution vehicle trajectory data is the foundation for many Intelligent Transportation System (ITS) applications. By providing fine-grained spatiotemporal information on vehicle positions and dynamics, such data enables the discovery and modeling of traffic flow patterns \cite{kanagaraj2015trajectory,li2020trajectory} and driving behaviors \cite{zhang2024bayesian}. Beyond offline analysis, trajectory data also plays a critical role in real-time ITS operations to improve safety and mobility, supporting applications such as traffic state estimation \cite{zhao2019various}, traffic signal optimization \cite{feng2015a, li2024a}, and proactive traffic safety management \cite{zhu2025vehicle}.

In practice, vehicle trajectories are primarily acquired through either probe vehicles or infrastructure-based sensing systems. Although probe vehicles generate spatially continuous trajectories, the resulting data are often sparse and contain only partial information due to limited market penetration rates. Infrastructure-based sensors, on the other hand, are designed for monitoring all vehicles within their detection zones, and generate ``ground-truth" traffic information. With the rapid development of sensing technologies, modern infrastructure platforms are increasingly integrated with deep-learning–based object detection algorithms \cite{jiang2022review, yin2021center}, enabling vehicle-level detection and tracking. The primary sensing modalities include vision-based cameras, LiDAR, and radar, each characterized by distinct trade-offs in range, resolution, cost, and environmental adaptability.

Among these sensing modalities, radar offers superior robustness and detection range than others. It estimates object distance by transmitting electromagnetic waves and analyzing reflected echoes. Infrastructure-based radar systems typically adopt frequency-modulated continuous wave (FMCW) architectures, in which range and velocity are jointly extracted through beat frequency analysis and Doppler processing \cite{patole2017automotive}. Owing to the longer wavelength, radar signals experience reduced atmospheric attenuation in conditions such as fog and rain, thereby maintaining reliable performance in adverse environments and enabling long-range detection. These characteristics make radar particularly suitable for continuous and long-range monitoring in highway and urban-arterial settings such as speed measurement \cite{kim2017assessing}, traffic counting \cite{tan2023bidirectional}, and vehicle detection \cite{lim2021lane}.
Nevertheless, the longer wavelength of the radar inherently limits angular resolution, resulting in coarse spatial representation and challenges in accurately estimating object shape and size. Radar measurements are also susceptible to multipath reflections, which can generate ghost detections and positional uncertainty. Moreover, conventional Doppler-centric processing frameworks prioritize moving targets and may exhibit reduced sensitivity to stationary or slow-moving objects. Collectively, these limitations pose significant challenges for reliable radar-based traffic monitoring.
Given both advantages and drawbacks of radar sensing, a key question in the deployment of radar for traffic monitoring lies in how to comprehensively evaluate detection and tracking performance in real-world settings. Existing studies primarily report aggregated traffic metrics, such as vehicle counts \cite{tan2023bidirectional}, or validate detection and tracking results against only a small number of selected vehicles \cite{lim2021lane}. A systematic and large-scale assessment of radar detection performance at multiple levels remains unexplored.

To address this gap, we introduce a dual-modality dataset of naturalistic vehicle trajectories, called \textbf{DRaT}, which stands for \textbf{D}rone and \textbf{Ra}dar \textbf{T}rajectories, designed to systematically evaluate radar-based detection and tracking performance against drone-derived ground truth. The dataset was collected at a highway merging area in Fort Worth, TX. During the data collection period, diverse and heterogeneous traffic flow patterns were observed. Radar performance is evaluated through three levels of aggregation, including: 1) individual vehicle detection; 2) trajectory tracking; and 3) macroscopic traffic parameter estimation. Meanwhile, we open-source the DRaT dataset on Zenodo to enable reproducible research on infrastructure-based sensing systems.

\section{Experiment Description}

\subsection{Study area}

\begin{figure}[t]
\centering
\includegraphics[width=0.9\columnwidth]{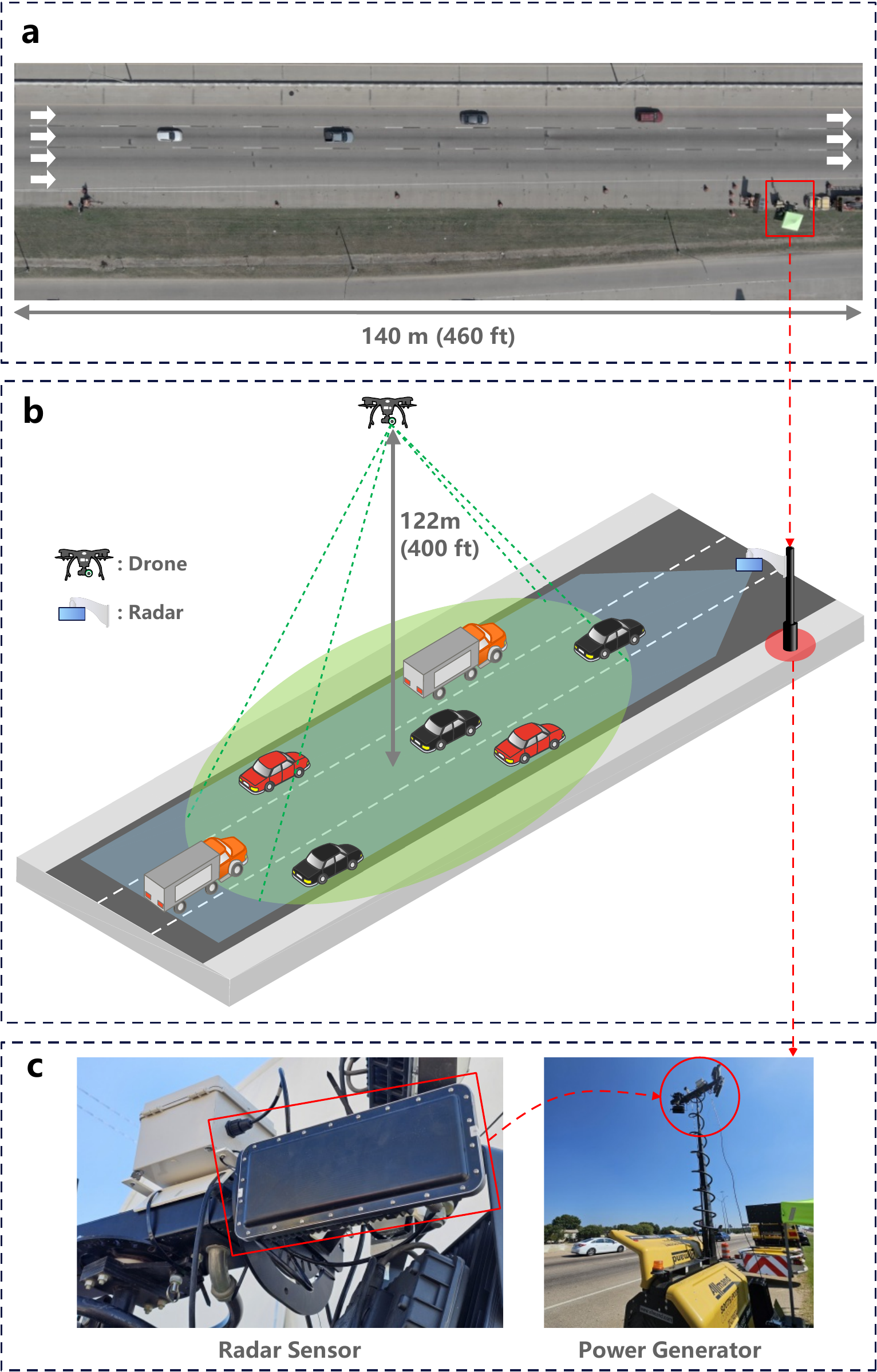}
\caption{{Study area and sensor setup for the experiment.}
\textbf{a.} Study area captured by drone.
\textbf{b.} Overview of sensor setup.
\textbf{c.} Radar mounting on a portable power generator.}
\label{fig:sensor_setup}
\end{figure}

The study area is located along the Ronald Reagan Memorial Highway in Fort Worth, Texas. As illustrated in Fig.~\ref{fig:sensor_setup}.a, the selected highway section spans approximately 140 m (460 ft) to align with the effective drone detection range. The analysis focuses exclusively on eastbound traffic, which moves from left to right in Fig.~\ref{fig:sensor_setup}.a. The roadway section consists of three regular lanes and one merging lane, forming a transition from four lanes to three within the segment. The data collection lasted approximately 80 minutes and captured two representative traffic conditions: a 43-minute free-flow period and a 37-minute congested period. During the free-flow condition, the average density was 40.79 veh/km, with a space-mean speed of 95.27 km/h and a traffic volume of 3886.12 veh/h. In contrast, the congested condition exhibited a higher density of 122.93 veh/km, a reduced space-mean speed of 28.70 km/h, and a traffic volume of 3527.43 veh/h. The two different traffic states help us comprehensively evaluate the radar's performance.

\subsection{Sensor specifications and setup}

\subsubsection{Radar}

The radar used in this study is a commercial product. It adopts an FMCW architecture and operates in the 24.25–24.50 GHz K-band. As summarized in Table~\ref{tab:radar_specs}, the radar provides a long detection range and a wide horizontal field-of-view (FOV), making it well-suited for traffic monitoring tasks. In addition, as shown in Fig.~\ref{fig:sensor_setup}.c, the radar was equipped with a large antenna aperture, providing sufficient angular resolution to support vehicle-dimension estimation.

\begin{table}[b]
\centering
\caption{Radar specifications}
\label{tab:radar_specs}
\small
\begin{tabular}{l c}
\hline
\textbf{Parameter} & \textbf{Value} \\
\hline
Operating frequency band & 24.25--24.5 GHz (K-band) \\
Detection range & 213--274 m (700--900 ft) \\
Horizontal field-of-view & 110$^\circ$ \\
Maximum angle to traffic flow & 60$^\circ$ \\
\hline
\end{tabular}
\end{table}

\subsubsection{Drone}
The drone selected for the experiment was also a commercial product. As shown in Table~\ref{tab:drone_specs}, the platform supports multi-constellation GNSS positioning and provides stable hovering with sub-meter accuracy. Videos can be recorded with a high resolution using a 3-axis stabilized gimbal, ensuring the clarity and consistency of the videos.

\begin{table}[b]
\centering
\caption{Drone specifications}
\label{tab:drone_specs}
\small
\begin{tabular}{l c}
\hline
\textbf{Parameter} & \textbf{Value} \\
\hline
Video resolution & Up to 5.4K p30, 4K p60 \\
GNSS support & GPS, GLONASS, Galileo \\
Maximum wind resistance & 10.7 m/s \\
Hovering accuracy & $\pm$0.1 m vertical, $\pm$0.3 m horizontal \\
Gimbal stabilization & 3-axis mechanical stabilization \\
\hline
\end{tabular}
\end{table}

An overview of the sensor configuration is presented in Fig.~\ref{fig:sensor_setup}.b. The drone hovered at an altitude of 122 m (400 ft), capturing 4K video at 30 fps from a bird’s eye view (BEV). As demonstrated in Fig.~\ref{fig:sensor_setup}.a, the right edge of the drone footage was aligned to include the radar location, maximizing the overlapping coverage area between the two sensors.

The radar sensor was mounted on the extendable pole of a portable power generator, as shown in Fig.~\ref{fig:sensor_setup}.c. The radar was installed at a height of approximately 5.5 m (18 ft) with a 15° downward tilt toward the traffic flow, balancing detection range, detection occlusions, and near-field blind regions. In addition, the radar’s GPS location and azimuth angle were measured manually to enable accurate georeferencing of detection results.
Since the drone provided an effective longitudinal roadway coverage of about 140 m (460 ft), radar detections were also limited to the same range for consistent cross-modality evaluation.

\section{Trajectory Extraction and Georeferencing}
\subsection{Drone}
Vehicle trajectories were extracted from the drone footage using the open-source Geo-trax framework \cite{fonod2025advanced}, which combines a pre-trained YOLO detector \cite{jocher2023ultralytics} with the BoT-SORT algorithm \cite{aharon2022bot} for vehicle detection and multi-object tracking. To mitigate the effects of drone hovering instability, each video frame was first stabilized by registering it to a selected master reference frame, thereby compensating for small frame-to-frame translations and rotations. The stabilized master frame was then registered to a pre-collected orthophoto with a ground sampling distance of approximately 0.059 m/pixel, enabling image coordinates to be transformed into real-world coordinates. The orthophoto was carefully georeferenced using multiple ground control points (GCPs), which provided the pixel-to-UTM coordinate transformation.

\begin{figure}[t]
\centering
\includegraphics[width=0.92\columnwidth]{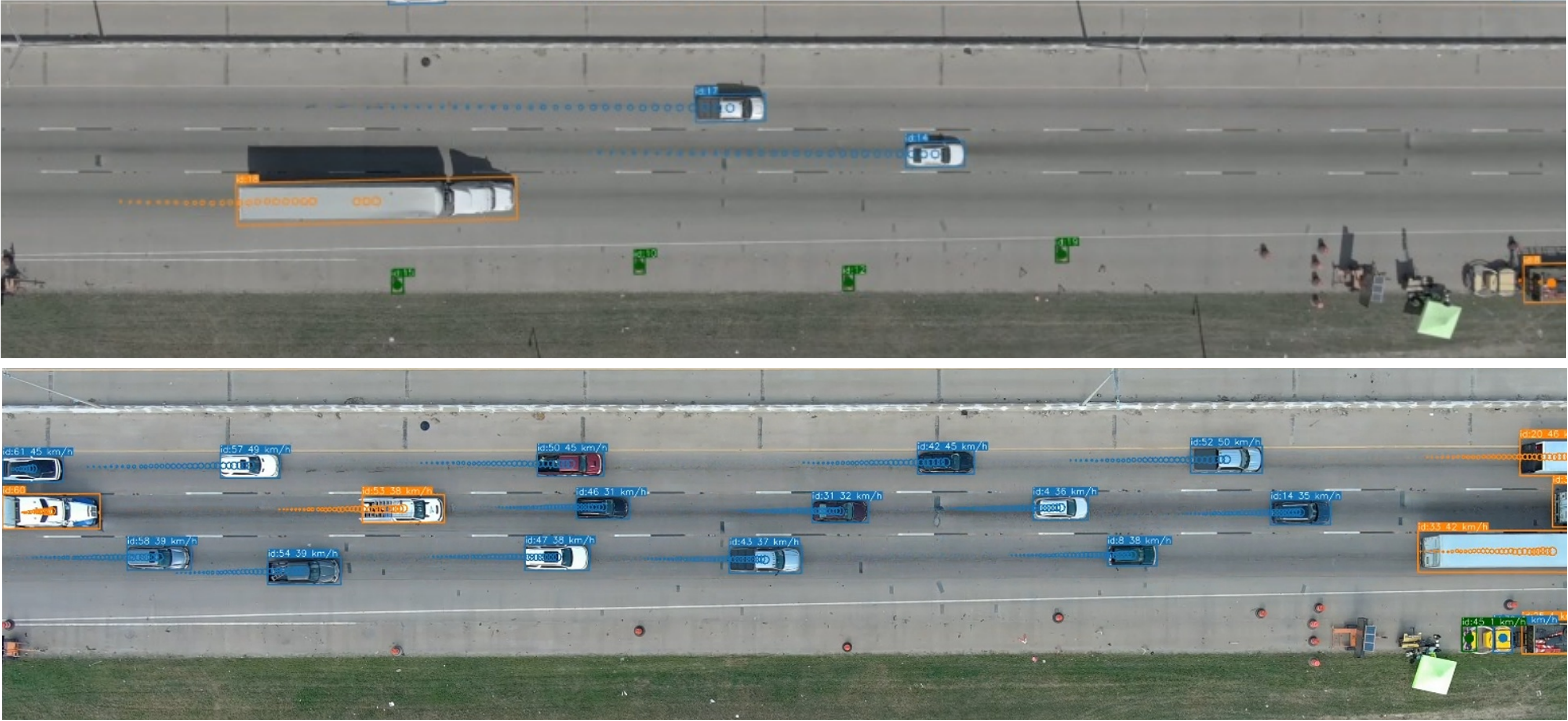}
\caption{Drone detection results.}
\label{fig:drone_detection}
\end{figure}

Because the orthophoto serves as the primary spatial reference with a consistent ground scale, the resulting trajectories are largely independent of drone GPS errors and hovering inaccuracies. Registering the stabilized imagery to georeferenced roadway features also helps reduce perspective-related distortions. In addition, roadway elevation variations within the selected study area are expected to have a negligible effect on the georeferenced trajectories. Therefore, the drone-derived ground truth can achieve high spatial consistency, with its accuracy mainly determined by the quality of video stabilization, orthophoto registration, and image-to-world transformation.

The drone-derived vehicle trajectories were refined by a per-frame greedy Non-Maximum Suppression (NMS) procedure \cite{gong2021review} that, scanning boxes in order of decreasing length, suppresses any later detection within 1.5\,m or with IoU $\geq 0.1$, or is $\geq 10\%$ engulfed by an already-kept box. The 30\,Hz drone trajectory was downsampled to the radar’s exact 8\,Hz timestamps through linear interpolation, yielding a one-to-one temporal correspondence with radar detections.
After post-processing, 5{,}278 vehicle trajectories were extracted from the drone footage, comprising 392{,}707 individual vehicle detections after radar-timestamp alignment. Examples of drone detection results are shown in Fig.~\ref{fig:drone_detection}. The high fidelity of drone-based detection and the robust georeferencing process support their use as ground truth in evaluating radar detection performance in the subsequent analysis.

\begin{figure}[t]
\centering
\includegraphics[width=0.99\columnwidth]{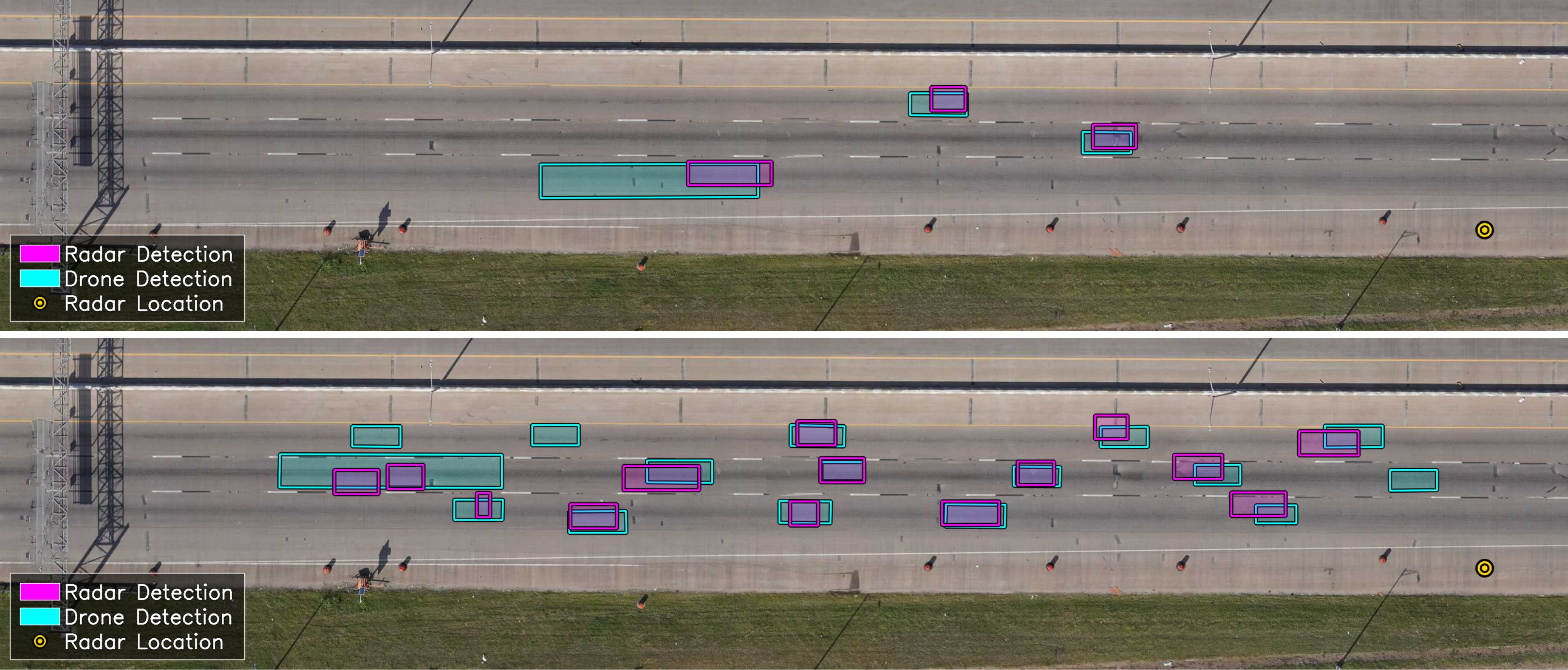}
\caption{Radar detection results compared to drone-derived ground truth.}
\label{fig:radar_detection}
\end{figure}

\subsection{Radar}
The radar sensor integrates a built-in object detection and tracking algorithm that directly outputs vehicle-level attributes, including positions, speeds, dimensions, and headings of detected objects with a resolution of 8 Hz. The radar outputs were refined using the same NMS procedure as drone data to eliminate redundant false-positive detections.
To ensure spatial consistency between the two sensing modalities, detected object positions were converted to GPS coordinates, using the measured GPS location and azimuth angle of the radar installation, following the same procedure in \cite{zhu2025smart}. 
To improve georeferencing accuracy, the radar reference position was measured using a Real-Time Kinematic (RTK)-corrected GPS receiver. The radar azimuth angle was initially measured with a compass and subsequently refined by aligning radar-derived trajectories with the road direction.

After post-processing, 5,430 vehicle trajectories were obtained, comprising 282,712 individual vehicle detections. Examples of radar detections under light and heavy traffic conditions are shown in Fig.~\ref{fig:radar_detection} (corresponding to Fig.~\ref{fig:drone_detection}), where they are overlaid on the orthophoto and compared with drone detections from the same timestamps. In the figure, radar-detected bounding boxes are shown as purple rectangles, drone detections are in cyan, and the radar sensor location is indicated by a yellow dot at the bottom right.

\section{Analysis and Evaluation}

\subsection{Individual vehicle detection}
To evaluate radar performance in detecting individual vehicles, a Hungarian-algorithm–based matching procedure \cite{zhu2025empowering} with an Intersection-over-Union (IoU) threshold of 0.1, was employed to associate radar detections with ground-truth detections on a frame-by-frame basis. Based on the matching results, detections were categorized as true positives (TPs), false positives (FPs), and false negatives (FNs), which were subsequently used to calculate performance metrics, including precision, recall, and F1 score.




\begin{table}[b]
\centering
\caption{Radar detection performance under different traffic conditions (IoU threshold = 0.1).}
\label{tab:individual_detection}
\small
\begin{tabular}{lccc}
\hline
\textbf{Traffic Condition} & \textbf{Precision} & \textbf{Recall} & \textbf{F1-score} \\
\hline
Free-flow   & 0.802 & 0.646 & 0.716 \\
Congested   & 0.775 & 0.538 & 0.635 \\
Overall     & 0.783 & 0.568 & 0.659 \\
\hline
\end{tabular}
\end{table}

\begin{figure}[t]
\centering
\includegraphics[width=0.90\columnwidth]{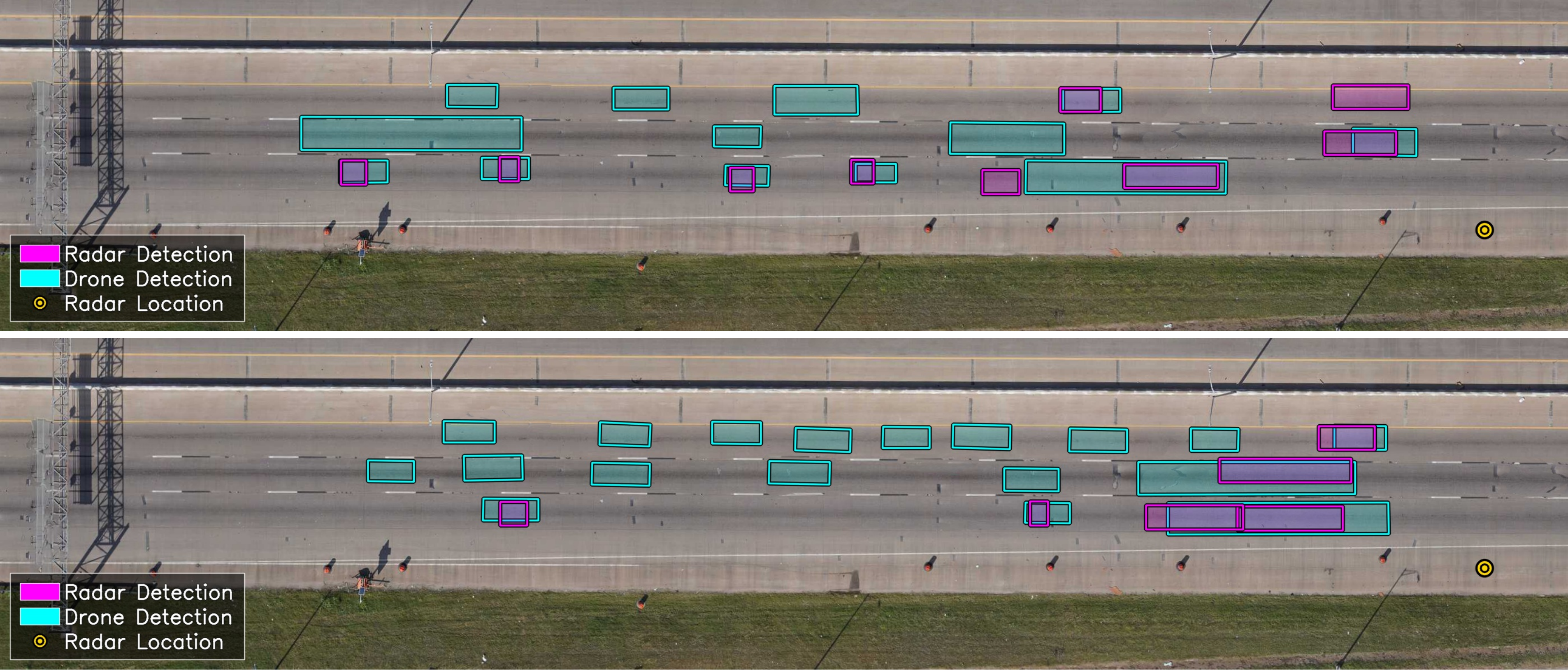}
\caption{Demonstration of occlusion in radar detection}
\label{fig:occlusion}
\end{figure}

As summarized in Table~\ref{tab:individual_detection}, the overall precision and recall of radar detection are 78.3\% and 56.8\%, respectively. This indicates that over 78\% of radar detections are deemed to be correct, while about 57\% of ground-truth vehicles in each frame are successfully detected.
A noticeable performance degradation is observed under congested traffic conditions, likely due to increased vehicle density and reduced speeds. Firstly, Doppler-centric radar data processing methods are inherently less sensitive to slow-moving targets, which are more prevalent during congestion. Moreover, in high-density scenarios, closely spaced vehicles and occlusions caused by larger vehicles restrict the radar’s line-of-sight coverage and target separation, leading to more missed detections (i.e., reduced recall). Fig.~\ref{fig:occlusion} provides two examples of occlusions caused by semi-trucks.

\begin{figure}[t]
\centering
\includegraphics[width=0.90\columnwidth]{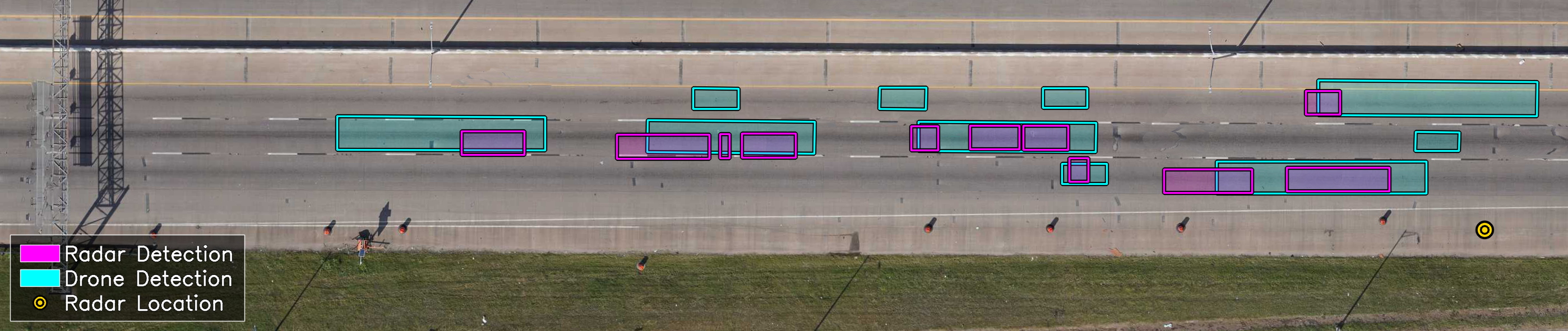}
\caption{Demonstration of radar's difficulty in estimating vehicle dimensions}
\label{fig:dimension}
\end{figure}

\begin{figure}[t]
\centering
\includegraphics[width=0.82\columnwidth]{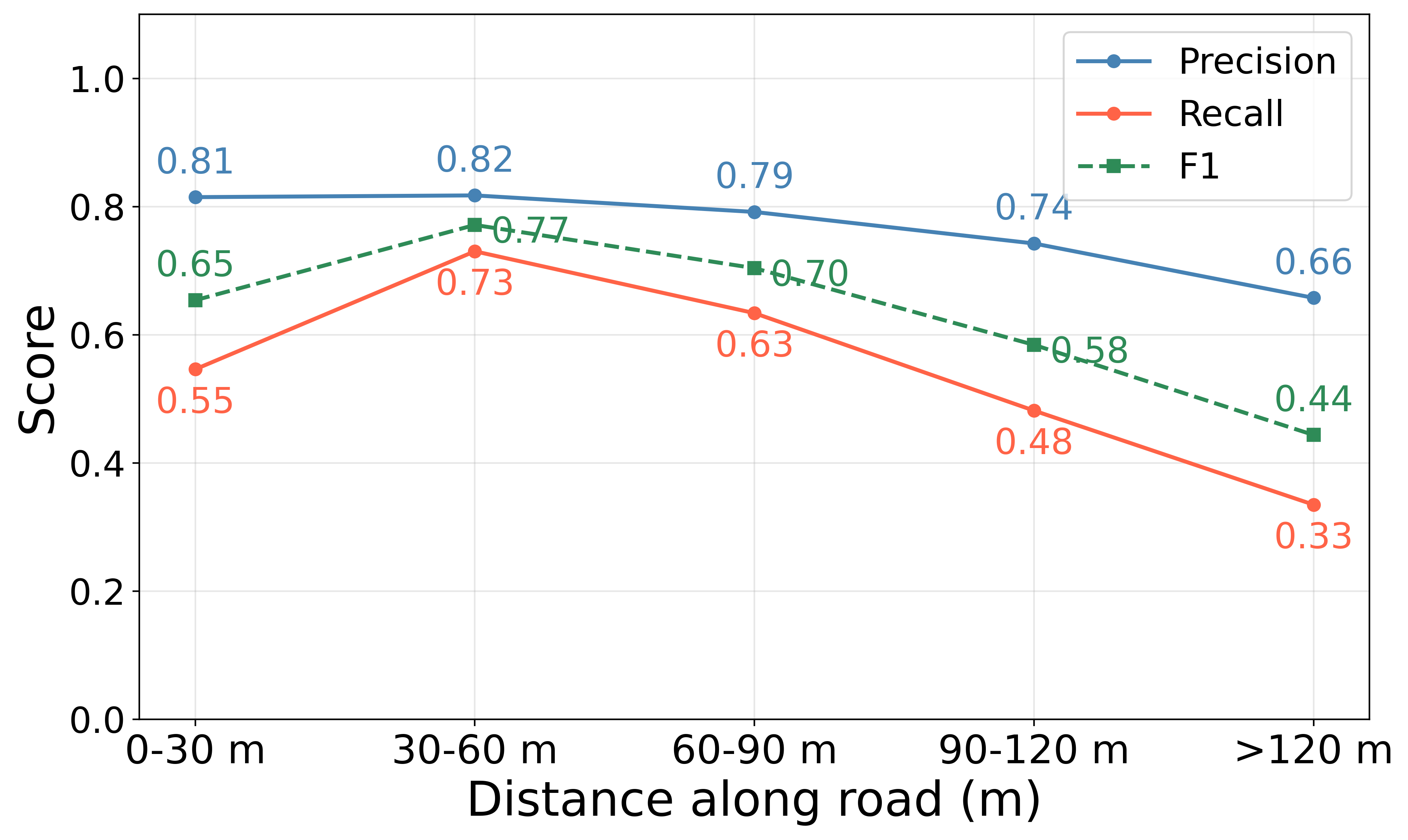}
\caption{Precision and recall within different longitudinal segments.}
\label{fig:distance_seg}
\end{figure}

At the same time, dense traffic conditions increase the likelihood of multipath reflections from adjacent vehicles, generating potential ghost detections. In addition, due to limited angular resolution, radar may fragment large vehicles (e.g., trucks) into multiple smaller detections, as illustrated in Fig.~\ref{fig:dimension}, thereby introducing additional false positives. These effects collectively contribute to reduced precision.

\begin{figure}[t]
\centering
\includegraphics[width=0.8\columnwidth]{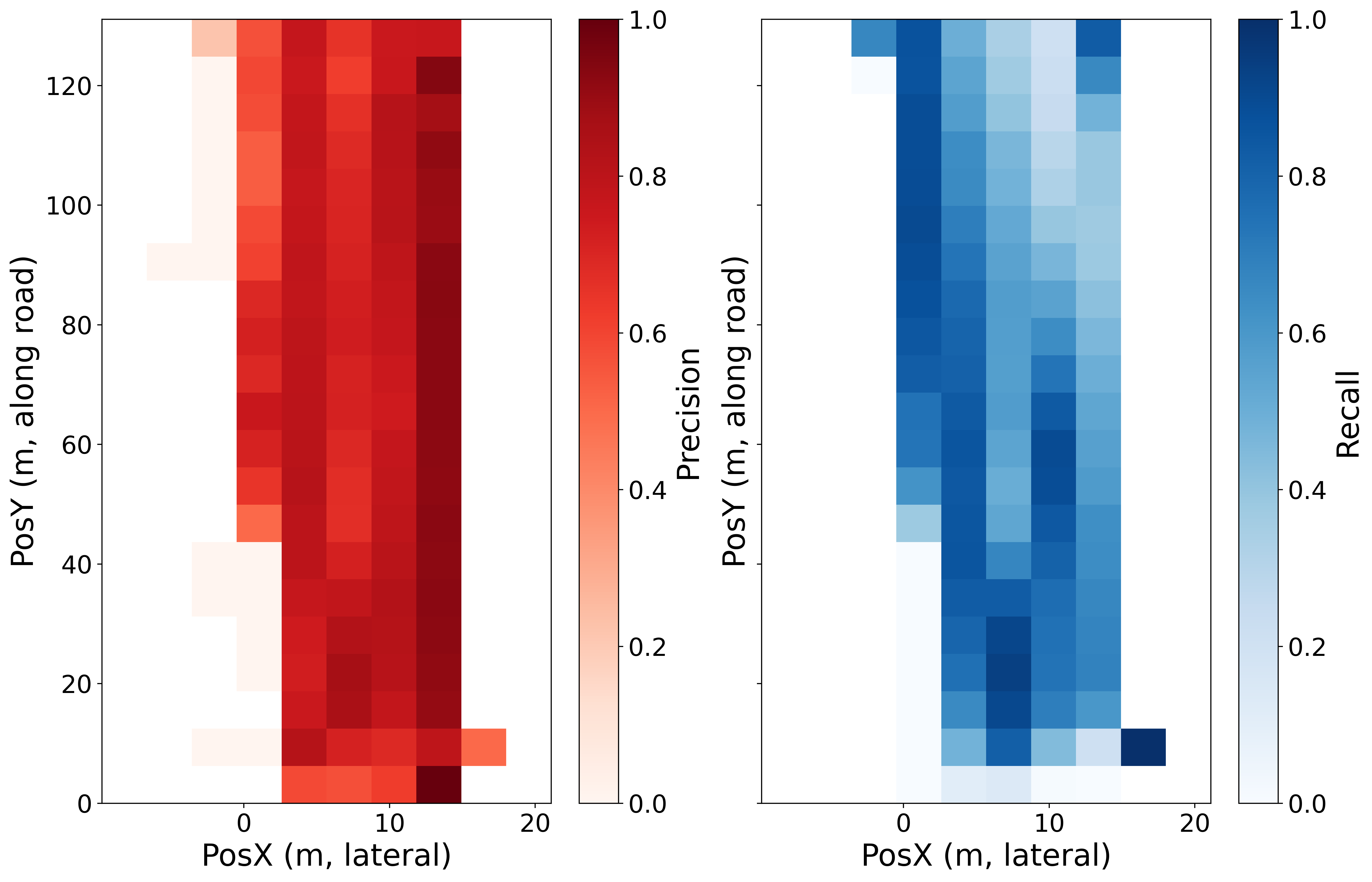}
\caption{Spatial distribution of precision and recall}
\label{fig:spatial_dist}
\end{figure}

From the distance perspective along the longitudinal direction of the roadway, both precision and recall peak in the 30-60\,m bin and decline at greater ranges, with precision dropping modestly from 0.82 to 0.66 and recall more sharply from 0.73 to 0.33, as shown in Fig.~\ref{fig:distance_seg}. Recall and F1 additionally fall inside 30\,m, producing a reversed-U trend absent in precision.
The lower near-field performance (Recall = 0.55) is mainly related to the radar’s downward boresight. With a 15$^\circ$ downward mounting angle, the center of the main-lobe footprint is projected onto the road surface at about 3.73 times the mounting height from the radar on the road surface. Vehicles close to the radar may fall outside the main high-gain region and are instead illuminated with reduced antenna gain, which lowers detection reliability. At longer ranges, performance decreases because of several combined effects. Signal strength weakens rapidly with distance, cross-range resolution becomes coarser, lower signal-to-noise ratio increases angular localization error, and vehicle occlusion becomes more severe, leading to erroneous detections.

The detailed spatial distribution of precision and recall is presented in Fig.~\ref{fig:spatial_dist}, where the radar is located at the origin and oriented along the positive vertical axis direction. Precision remains relatively stable across most longitudinal positions, with consistently high values in the central detection region and slight degradation near the boundaries, consistent with the trends observed in Fig.~\ref{fig:distance_seg}. In contrast, recall exhibits a more pronounced spatial variation, gradually decreasing toward the outer lanes, at farther distances, and within the near-field region close to the radar.

\begin{figure}[t]
\centering
\includegraphics[width=0.85\columnwidth]{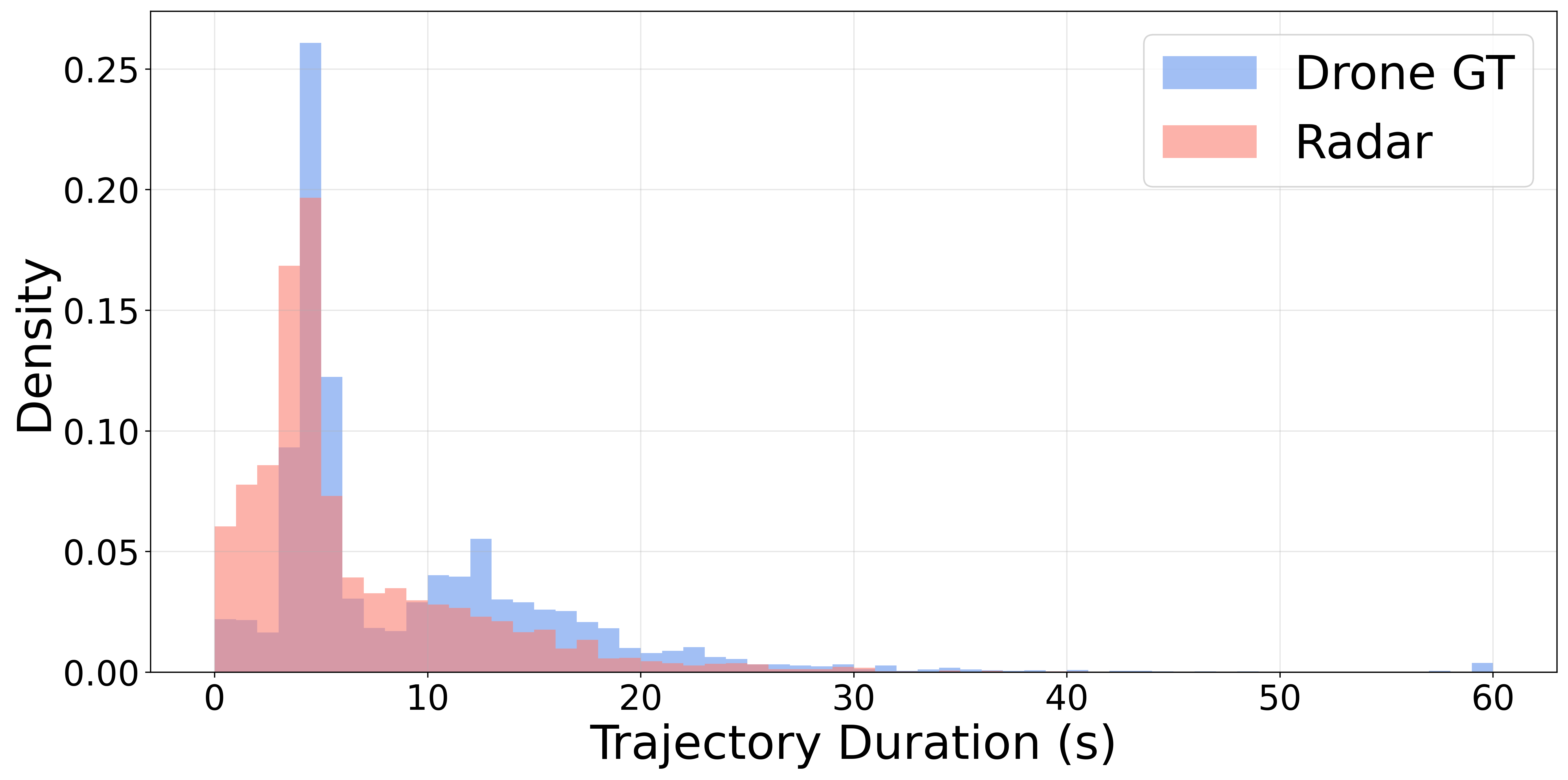}
\caption{Trajectory duration of radar detection}
\label{fig:trajectory_duration}
\end{figure}

\begin{table*}[htbp]
\centering
\caption{Radar trajectory tracking performance under different traffic conditions.}
\label{tab:tracking_performance}
\small
\setlength{\tabcolsep}{6pt}
\begin{tabular}{lccccc}
\hline
\textbf{Traffic Condition} &
\textbf{MOTA} &
\textbf{MOTP} &
\textbf{IDSW} &
\textbf{MT / PT / ML (\%)} &
\textbf{IDF1 (IDP / IDR)} \\
\hline

Overall &
0.535 &
1.79 m &
1,098 &
47 / 38 / 14 &
0.699 (0.83 / 0.60) \\

Free-flow &
0.601 &
1.79 m &
309 &
55 / 34 / 11 &
0.766 (0.86 / 0.69) \\

Congested &
0.509 &
1.79 m &
789 &
38 / 44 / 18 &
0.672 (0.82 / 0.57) \\

\hline
\end{tabular}
\end{table*}

\subsection{Trajectory tracking}
In this section, radar performance is further evaluated at the trajectory level. Fig.~\ref{fig:trajectory_duration} compares the duration of each tracked vehicle trajectory derived from radar and drone detections. The radar detection duration distribution is noticeably shifted toward the left relative to the ground-truth (drone) distribution, indicating that radar-derived trajectories tend to be shorter. This discrepancy is likely attributable to more missed detections at longer distance as shown in Fig. \ref{fig:spatial_dist}.

To evaluate radar's tracking performance, radar detections are matched with drone-derived ground truth using the same criterion as the individual detection evaluation. As summarized in Table~\ref{tab:tracking_performance}, we report Multiple Object Tracking Accuracy (MOTA), Multiple Object Tracking Precision (MOTP), identity switches (IDSW), the proportions of mostly tracked, partially tracked, and mostly lost trajectories (MT/PT/ML), and identity F1 score (IDF1), together with identity precision (IDP) and identity recall (IDR).

Overall, the radar achieves a MOTA of 0.535 and an IDF1 of 0.699, with an identity precision of 0.83 and identity recall of 0.60, indicating that radar-derived vehicle identities are generally reliable when successfully established, while incomplete trajectory coverage remains a major source of tracking error. In particular, 47\% of the ground-truth trajectories are mostly tracked, whereas 38\% are partially tracked and 14\% are mostly lost. Tracking performance deteriorates noticeably under congested conditions: MOTA decreases from 0.601 to 0.509, IDF1 decreases from 0.766 to 0.672, and the proportion of mostly tracked trajectories decreases from 55\% to 38\%. Meanwhile, identity switches increase from 309 in free flow to 789 under congestion, despite the smaller number of vehicles in the congested period. These results suggest that dense traffic primarily challenges the radar's ability to maintain continuous and consistent vehicle trajectories, likely due to increased occlusion and reduced target separation. In contrast, MOTP remains unchanged at 1.79~m across traffic conditions, indicating that the localization accuracy of successfully matched detections is relatively insensitive to traffic state.

\subsection{Macroscopic traffic parameter estimation}

\begin{table}[b]
\centering
\caption{Radar estimation errors for macroscopic traffic flow parameters.}
\label{tab:radar_macro_metrics}
\small
\begin{tabular}{lccc}
\hline
\textbf{Metric} & \textbf{MAE} & \textbf{RMSE} & \textbf{MAPE (\%)} \\
\hline
Density (veh/km) & 21.27 & 31.10 & 23.40 \\
Space-Mean Speed (km/h) & 1.37 & 1.85 & 3.76 \\
Volume (veh/h) & 829.43 & 918.18 & 23.83 \\
\hline
\end{tabular}
\end{table}

\begin{figure}[t]
\centering
\includegraphics[width=0.95\columnwidth]{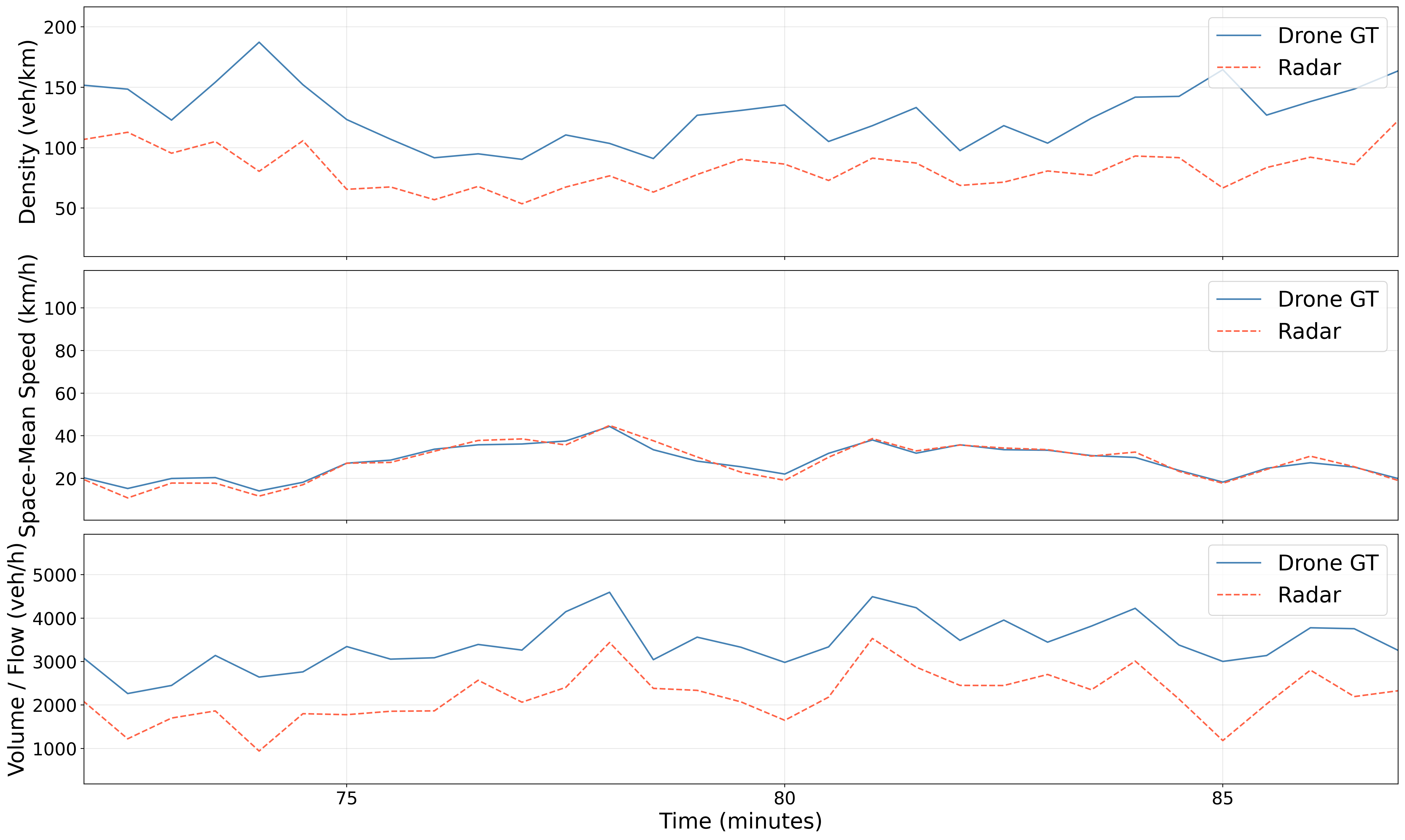}
\caption{Radar performance in macroscopic traffic metric estimation}
\label{fig:traffic_metrics}
\end{figure}

Finally, the radar performance is evaluated in estimating macroscopic traffic parameters including density, space-mean speed, and flow, using a 30-second time window.

As shown in Table~\ref{tab:radar_macro_metrics}, the radar achieves high accuracy in estimating space-mean speed, with an average MAE of only 1.37 km/h and a mean absolute percentage error (MAPE) below 4\%. In contrast, the estimates of density and volume deviate approximately 23\% from the ground truth in terms of MAPE. The relatively small difference between root mean squared error (RMSE) and MAE for these metrics suggests that the errors are fairly consistent over time, rather than being dominated by occasional large deviations. The primary source of error in density and volume estimation is the high missed detection rate within the observation area, which leads to systematic underestimation of traffic flow.
Fig.~\ref{fig:traffic_metrics} illustrates the parameter estimation results over a 15 mins window, providing a more intuitive visualization of the radar’s performance in capturing traffic dynamics.

\section{Discussions and Conclusions}


In this study, we introduced DRaT, a dual-modality dataset of naturalistic vehicle trajectories designed for the systematic evaluation of infrastructure-based radar sensing using drone-derived trajectories as ground truth. Meanwhile, DRaT supports reproducible research on infrastructure-based sensing systems, particularly in merging areas where lane-changing interactions and dense vehicle maneuvers are common. By providing the ground-truth trajectories, DRaT can also facilitate the development and evaluation of methods for identifying, characterizing, and potentially mitigating detection errors in infrastructure-based sensing systems.

The experimental results provide several practical insights for deploying radar sensing systems in ITS applications. Radar demonstrates effective vehicle detection and tracking capabilities, although its performance varies across traffic conditions and sensing ranges. In particular, detection performance degrades under congested traffic conditions and at longer sensing distances, while tracking performance also deteriorates under congestion due to increased occlusion and reduced target separation. These findings suggest that roadside radar is well suited for motion-centric monitoring and near-to-mid-range traffic sensing, but should be used with caution as a standalone modality for applications requiring complete vehicle detection and accurate traffic counting, particularly at longer ranges.

The identified error characteristics also provide guidance for multi-sensor fusion system design. Radar can provide reliable velocity measurements and robust perception under challenging weather and lighting conditions, making it valuable for motion-related traffic monitoring. Camera-based sensing can complement radar by offering higher spatial resolution and richer semantic information. Therefore, radar-vision fusion frameworks can adopt task-dependent weighting strategies: radar measurements may be assigned higher confidence for speed-related states, while vision-based measurements may receive greater weight for vehicle class and dimension estimation, especially in dense traffic scenarios where radar resolution is limited.

Several limitations of this study should be acknowledged. First, the radar used is a commercial off-the-shelf system with a built-in detection algorithm, which prevents examination on how alternative signal processing or detection algorithms might influence performance. Second, the results are based on a specific mounting height, different installation configurations may affect detection accuracy and coverage. Third, the evaluation is conducted under one highway segment, and additional road types (e.g., curves, ramps, or complex interchanges) should be considered to further assess the generalizability of the findings.

\section{Acknowledgment}
This research is supported by U.S. Department of Transportation through award \#69A3552541003. The views presented in this paper are those of the authors alone.

\bibliographystyle{IEEEtran}
\bibliography{refs}

\end{document}